\documentclass[preprint,12pt,nopreprintline]{elsarticle}

\usepackage{amsmath,amssymb}
\usepackage{booktabs}
\usepackage{graphicx}
\usepackage{multirow}
\usepackage{tabularx}
\usepackage{threeparttable}
\usepackage[table]{xcolor}
\usepackage{float}
\usepackage{placeins}
\usepackage{tikz}
\usetikzlibrary{arrows.meta,positioning,fit,calc}
\usepackage{microtype}
\usepackage{enumitem}
\usepackage{url}
\usepackage[unicode,pdfencoding=auto]{hyperref}


\definecolor{semanticblue}{HTML}{2F5D8C}
\definecolor{affectorange}{HTML}{D97732}
\definecolor{oursgreen}{HTML}{2F8F72}
\definecolor{softgray}{HTML}{F2F4F7}
\definecolor{tableblue}{HTML}{E8F2FA}
\definecolor{tableyellow}{HTML}{FFF4CC}
\definecolor{oursrow}{HTML}{FFF9E6}
\definecolor{semanticcell}{HTML}{D9EBF7}
\definecolor{affectcell}{HTML}{FFE9A8}
\hypersetup{colorlinks=true,linkcolor=semanticblue,citecolor=semanticblue,urlcolor=semanticblue}

\begin{document}

\begin{frontmatter}

\title{Semantic-First: Temporal Scheduling in Flow Matching for Reliable Emotional Image Generation}

\author{Zhenxuan Wang\corref{cor1}}
\ead{wzhenxuan2025@lzu.edu.cn}
\cortext[cor1]{Corresponding author.}
\address{Lanzhou University}

\begin{abstract}
Affect-controlled image generation should change how a scene feels while preserving what the prompt specifies. We study the temporal allocation of stochastic exploration and policy optimization in flow-matching models as a means of controlling this balance. During RL training, our \emph{semantic-first} schedule uses deterministic ordinary differential equation (ODE) sampling during the high-noise phase and confines stochastic differential equation (SDE) exploration and group relative policy optimization (GRPO) to a late window. Controlled experiments separate exploration location, optimization location, and terminal-step sampling, showing distinct semantic benefits from late exploration and late updates. On FLUX.1-dev, semantic-first improves prompt faithfulness, independently evaluated affect accuracy, and human-rated semantic preservation and overall preference. The expanded semantic audit records 2.7\% subject drift, while GRPO training time decreases by 41.1\% relative to original full-path Flow-GRPO. Window-width experiments characterize the affect--semantic and computational trade-offs, and replication on SD3.5-Medium reproduces the scheduling advantage. These results establish denoising time as a practical control dimension for reliable emotional image generation.

\end{abstract}


\end{frontmatter}

\section{Introduction}
\label{sec:introduction}

Emotion-aware image generation is useful in advertising, interface design, educational media, entertainment, and human--computer interaction, where the same semantic scene may need to convey different affective states. A practical system should therefore answer two questions simultaneously: \emph{what} should appear, and \emph{how} should it feel. The continuous valence--arousal (V--A) model provides a compact control space for the second question \cite{russell1980circumplex,gunes2011continuous}. EmotiCrafter further demonstrates that an emotion injection transformer (EIT) can condition text-to-image generation on continuous V--A targets \cite{emoticrafter2025}. Nevertheless, stronger affective modulation may change a prompt's main object, layout, or relations. In engineering deployment, a cheerful tree becoming fireworks is not a harmless stylistic variation; it is a specification failure.

Online reinforcement learning (RL) offers a direct way to optimize non-differentiable affect and semantic rewards. Flow-GRPO formulates stochastic sampling in flow-matching models as a Markov decision process and optimizes the full trajectory \cite{flowgrpo2025}. Its full-path SDE rollout creates exploration but is computationally expensive. MixGRPO reduces this overhead by using SDE sampling and GRPO optimization only inside a short window while using deterministic ODE sampling elsewhere \cite{mixgrpo2025}. In the original formulation, the window is mainly an efficiency and curriculum mechanism. Yet denoising time is not attribute-neutral: high-noise steps establish global scene structure, whereas low-noise steps increasingly affect local appearance and detail \cite{mixgrpo2025,esser2024flux}.

This observation motivates our central question: \emph{when should stochastic exploration be introduced if semantics and affect must share one probability-flow trajectory?} We compare two temporal orders under the same backbone, rewards, number of sampling steps, and SDE-window size. \emph{Emotion-first} exposes the high-noise phase to SDE exploration and later returns to an ODE trajectory. \emph{Semantic-first} keeps the high-noise phase deterministic and introduces stochastic exploration only after the scene structure has formed. In our implementation, step~0 is the highest-noise step, so semantic-first corresponds to a late SDE window $[20,24)$, not an early one.

\begin{figure}[!htb]
\centering
\includegraphics[width=\textwidth]{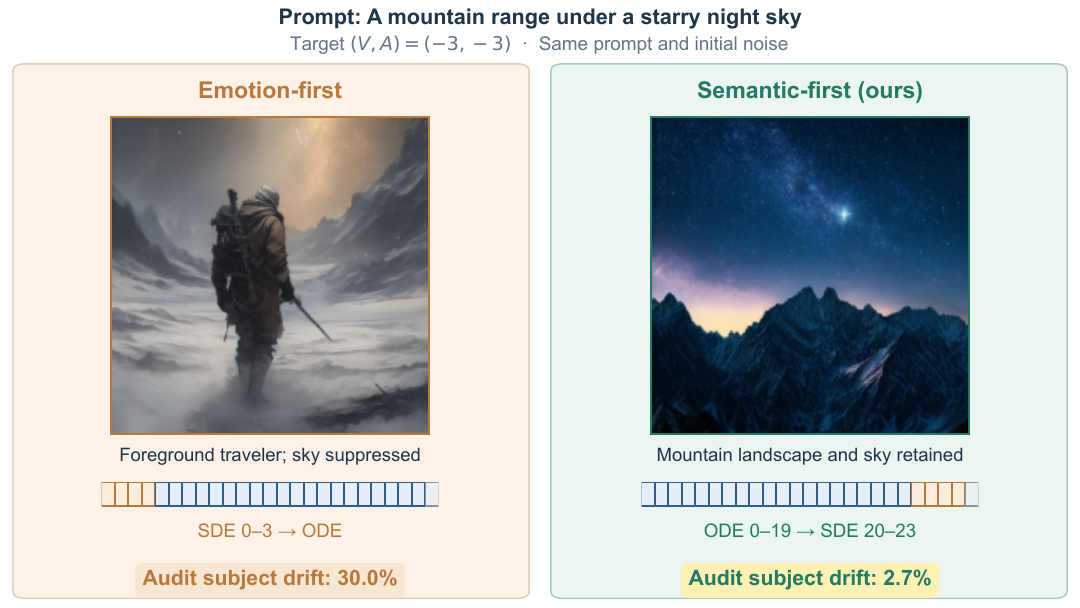}
\caption{With the same prompt, V--A target and initial noise, early exploration introduces a foreground traveler; late exploration retains the mountain landscape and night sky. Percentages are aggregate drift rates from the 150-image-per-method audit (Table~\ref{tab:order}). Vector-layout code: OpenAI Codex (desktop 26.1021.3705.0), reviewed by the author.}
\label{fig:motivation}
\end{figure}

On FLUX, full-path Flow-GRPO achieves the lowest V--A error under the training regressors, while semantic-first improves prompt alignment, cross-affect consistency, and independently evaluated affect accuracy. The proposed late-window schedule improves affect control over a non-RL baseline, improves the reported main metrics over progressive MixGRPO, and cuts end-to-end training cost. Controlled experiments separate where stochastic exploration occurs from where policy updates are applied, and a second flow-matching backbone tests the transfer of the scheduling principle. Figure~\ref{fig:motivation} illustrates the central goal: affective appearance should change while the prompt-defined scene remains recognizable.

Our contribution is an empirical study of temporal allocation: where stochastic exploration and policy optimization act along denoising time to improve affect control while preserving prompt semantics. We reuse the GRPO objective and test the scheduling effects through separate sampling-window and optimization-window interventions.

Our contributions are:

\begin{itemize}[leftmargin=*]
    \item We study the temporal placement of stochastic exploration and policy optimization as control variables for the affect--semantic trade-off.
    \item We instantiate a semantic-first schedule: ODE sampling at high noise followed by a fixed four-step SDE window at low noise, reusing the V--A EIT and MixGRPO objective.
    \item We evaluate the scheduling principle with separate sampling and optimization controls, four window widths, two flow-matching backbones, independent affect evaluation, TIFA, semantic auditing, efficiency measurements, and a 30-participant user study.
    \item We establish a practical operating point that combines strong affect control, prompt-content preservation, and lower training cost, supported by paired comparisons across automatic and human evaluations.
\end{itemize}

\section{Related work}
\label{sec:related}

\subsection{Continuous emotional image generation}

Visual affect has been modeled with discrete categories and continuous dimensions, with valence and arousal providing a widely used two-dimensional representation \cite{russell1980circumplex,gunes2011continuous}. Earlier generation and editing approaches used emotional color transfer, semantic emotion concepts, or diffusion-based image manipulation \cite{liu2018color,emogen2024,emoedit2024}. EmotiCrafter learns a continuous emotion embedding and injects it into prompt features, allowing one prompt to be rendered at different V--A coordinates \cite{emoticrafter2025}. Anchor-regularized GRPO combines continuous V--A conditioning with a neutral-image semantic reference to balance affective alignment and content preservation \cite{dang2026vaanchoring}. We study how the temporal allocation of exploration and policy updates controls the balance between affect alignment and prompt-content preservation.

\subsection{Flow matching and mixed ODE--SDE sampling}

Diffusion and flow models admit deterministic probability-flow ODEs and related stochastic sampling processes \cite{song2021sde,lipman2023flow,albergo2023stochastic}. Rectified flow provides an efficient generative formulation \cite{liu2023rectified}, and transformer-based flow models such as SD3 \cite{esser2024flux} and FLUX \cite{flux2024} support high-resolution text-to-image synthesis. Deterministic solvers accelerate generation \cite{song2021ddim,lu2022dpmsolver}, whereas stochastic transitions provide the action distribution needed by online policy optimization.

MixGRPO combines these modes by placing SDE transitions inside a sliding window and ODE transitions outside it \cite{mixgrpo2025}. Its progressive schedule moves from low signal-to-noise ratio (SNR) to high SNR. Recent work further shows that flow-model policy optimization is temporally non-uniform. TempFlow-GRPO concentrates branching at selected timesteps and applies noise-aware optimization weights \cite{tempflowgrpo2026}; Stepwise-Flow-GRPO assigns credit from estimated reward improvements at individual transitions \cite{savani2026stepwise}; and BranchGRPO reuses shared trajectory prefixes to improve rollout stability and efficiency \cite{branchgrpo2026}. Neighbor GRPO instead avoids SDE conversion by constructing a contrastive policy objective around neighboring deterministic ODE trajectories \cite{neighborgrpo2026}. These methods target temporal credit assignment, rollout reuse, or alternative policy construction. In contrast, we keep the reward, backbone, and four-transition budget fixed and treat the \emph{location} of stochasticity as an attribute-control variable. We examine how the temporal placement of exploration and policy optimization changes affect accuracy and prompt-content preservation.

The relation between noise level and structural preservation also motivates SDEdit, which adds noise to an image guide and denoises it to balance input faithfulness and realism \cite{meng2022sdedit}. In video generation, FreeNoise reschedules initial noise across frames and uses windowed temporal attention to maintain long-range content consistency \cite{qiu2024freenoise}. These studies provide context for time-dependent control: SDEdit selects an editing noise level, FreeNoise manages video-frame coherence, and our method allocates exploration and reward optimization across denoising transitions in affect-conditioned image generation.

\subsection{Reinforcement learning for image generation}

Reward-weighted fine-tuning and policy-gradient methods have aligned diffusion models with human preferences and task-specific objectives \cite{black2023ddpo,fan2023dpok,wallace2024diffusiondpo}. GRPO normalizes rewards within groups and avoids a learned value network \cite{shao2024deepseekmath}. Flow-GRPO and DanceGRPO adapt this idea to flow-based visual generation \cite{flowgrpo2025,dancegrpo2025}. Because reward optimization may exploit imperfections in evaluators, multi-objective evaluation and semantic auditing remain essential. Our experiments use separate affect and semantic rewards, report both outcome families, and include human evaluation rather than collapsing them into a single score.

Application-oriented RL spans control, recommendation, and energy-management tasks, as reviewed by Sivamayil et al.\ \cite{sivamayil2023rl}. Beyond image synthesis, PACE-MoE coordinates adaptive domain sampling and domain-specific LoRA for medical visual grounding, retaining the GRPO objective while changing the allocation of data and parameter updates \cite{wang2026pacemoe}. Our choice of GRPO follows the structure of image generation: terminal image rewards rank a group of stochastic trajectories, and within-group normalization supplies a policy-gradient signal without a learned value network. Generation design also depends on the target output. SignCraft combines speech-to-sign-gloss processing with gesture synthesis for consumer devices \cite{rajasekar2026signcraft}; our task instead preserves text-specified scene content while varying continuous image affect. These comparisons situate the policy choice and generation objective in their respective application settings.

\section{Methodology}
\label{sec:method}

\subsection{Problem formulation}

Let $c$ denote a neutral text prompt, $e=(v,a)\in[-3,3]^2$ a target valence--arousal coordinate, and $x_T$ the generated image. We seek a policy that reduces affect error while preserving the prompt-conditioned subject and incurring practical training cost. For a schedule $W$, the engineering objective is naturally multi-objective:
\begin{equation}
 W^{\star}\in\operatorname{ParetoArgMin}_{W}
 \left(E_{\mathrm{VA}}(W),-S_{\mathrm{sem}}(W),C_{\mathrm{train}}(W)\right),
 \label{eq:pareto}
\end{equation}
where $E_{\mathrm{VA}}$ is affect error, $S_{\mathrm{sem}}$ measures semantic fidelity or consistency, and $C_{\mathrm{train}}$ is computational cost.

\subsection{V--A emotion injection}

We adapt the residual EIT of EmotiCrafter \cite{emoticrafter2025} to FLUX.1-dev. Separate MLPs map valence and arousal to token sets $Z_v$ and $Z_a$; the adapter combines them with frozen neutral-prompt features $H_c$ to predict a residual:
\begin{equation}
 \Delta H=\mathrm{EIT}(H_c,Z_v,Z_a),
 \qquad H_{c,e}=H_c+\Delta H.
 \label{eq:eit}
\end{equation}
The FLUX-compatible adapter first projects the 4096-dimensional T5 tokens to width 768. Each scalar V/A input is mapped to eight 768-dimensional tokens. Four blocks apply self-attention, valence cross-attention, arousal cross-attention, and a feed-forward network, using 12 attention heads. A final normalization and $768\rightarrow4096$ projection produce the residual in the original T5 conditioning space; the maximum prompt length is 128. The unchanged pooled CLIP embedding supplies the separate global text condition. During adapter pretraining only, $\alpha=1.5$ scales the target residual,
\begin{equation}
 H^{\star}_{c,e}=H_c+\alpha(H_e-H_c),
 \label{eq:eit-target}
\end{equation}
where $H_e$ is the frozen T5 embedding of the paired emotional rewrite. During GRPO, the frozen text encoders and VA-EIT supply cached $H_{c,e}$, the unchanged pooled CLIP condition, and FLUX text identifiers.

\subsection{Mixed sampling with a fixed stochastic window}

We use 25 sampling transitions indexed by $k\in\{0,\ldots,24\}$, where $k=0$ is the highest-noise transition. The terminal transition $k=24$ is excluded from the policy objective, so GRPO-optimized indices are 0--23. For a four-step stochastic window beginning at $l$,
\begin{equation}
 W(l)=\{l,l+1,l+2,l+3\}, \qquad 0\le l\le20.
 \label{eq:window}
\end{equation}
The transition is stochastic only within $W(l)$:
\begin{equation}
 x_{k+1}=\begin{cases}
 \mu_{\theta}(x_k,t_k,H_{c,e})+\sigma_k\epsilon_k,
     & k\in W(l),\quad \epsilon_k\sim\mathcal{N}(0,I),\\
 \Phi_{\theta}(x_k,t_k,H_{c,e}), & k\notin W(l),
 \end{cases}
 \label{eq:mixed}
\end{equation}
where $\Phi_{\theta}$ is the deterministic ODE update and $\mu_{\theta},\sigma_k$ define the SDE transition inherited from Flow-GRPO/MixGRPO \cite{flowgrpo2025,mixgrpo2025}. Policy log-probabilities and GRPO gradients are computed only for $k\in W(l)$. For the controlled experiments, we distinguish the stochastic-transition set $S$ from the optimized-transition set $U\subseteq S$. Equation~\eqref{eq:mixed} uses $S$ in place of $W(l)$, while Eq.~\eqref{eq:grpo} averages the policy loss over $U$. The proposed schedule sets $S=U=W(20)$; separating them permits matched interventions on exploration and optimization.

\begin{figure}[!htb]
\centering
\includegraphics[width=\textwidth]{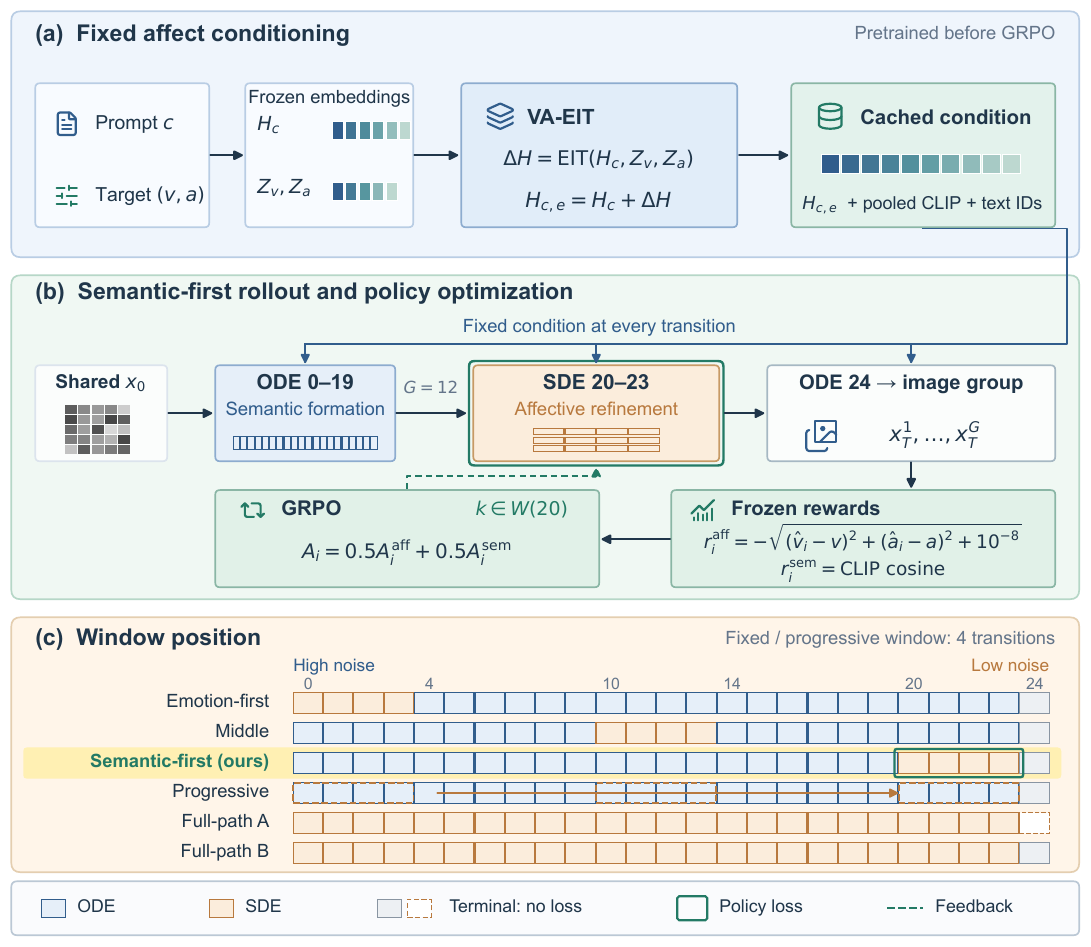}
\caption{Semantic-first training. (a) The residual EIT maps neutral-prompt T5 features and a continuous V--A target to fixed FLUX conditions. (b) ODE steps 0--19 precede SDE exploration and GRPO loss at 20--23, followed by ODE readout at 24. The trained policy serves all inference steps. (c) Fixed/progressive schedules use four stochastic transitions; A uses an SDE terminal step and B uses ODE. Controls: Table~\ref{tab:schedule_config}. Vector-layout code: OpenAI Codex (desktop 26.1021.3705.0), reviewed by the author.}
\label{fig:method}
\end{figure}

Figure~\ref{fig:method} summarizes the training pipeline. Semantic-first uses ODE transitions 0--19 for the high-noise phase, then allocates affect-conditioned SDE exploration to transitions 20--23. Affect and semantic rewards update the policy within this late window. Mixed schedules use an ODE terminal readout at step 24; the original full-path baseline A uses SDE, while matched baseline B and all sampling/optimization controls use ODE. This convention separates terminal sampling from window placement.

\label{sec:schedules}
Fixed-width comparisons use $w=4$: Emotion-first selects $W(0)$, Middle selects $W(10)$, and Semantic-first selects $W(20)$. With transition 24 reserved for ODE readout, 20--23 is the latest eligible four-transition window; this width matches the early/middle comparison budget, and the width study in Section~\ref{sec:width} characterizes its balance of affect control, semantic preservation, and cost. Progressive MixGRPO moves the window from $l=0$ to $l=20$. Full-path A samples all 25 transitions stochastically; B uses ODE at 24. Both optimize indices 0--23. Table~\ref{tab:schedule_config} specifies the separate exploration and optimization controls.

\subsection{Joint GRPO objective}

For each prompt and target, $G=12$ images are generated from the same initial noise and diverge through their active SDE transitions. The affect reward is the negative Euclidean distance predicted by frozen CLIP image regressors:
\begin{equation}
 r_i^{\mathrm{aff}}=-\sqrt{(\hat v_i-v)^2+(\hat a_i-a)^2+10^{-8}}.
 \label{eq:vareward}
\end{equation}
The semantic reward $r_i^{\mathrm{sem}}$ is CLIP text--image cosine similarity \cite{radford2021clip,hessel2021clipscore}. Each reward is normalized within its prompt group,
\begin{equation}
 A_i^q=\frac{r_i^q-\mu_q}{s_q+10^{-8}}, \qquad
 A_i=0.5A_i^{\mathrm{aff}}+0.5A_i^{\mathrm{sem}},
 \label{eq:advantage}
\end{equation}
where $q\in\{\mathrm{aff},\mathrm{sem}\}$. For transition $k$ in the active window, the policy ratio is
\begin{equation}
\rho_{i,k}=\frac{\pi_{\theta}(x_{i,k+1}\mid x_{i,k})}
{\pi_{\theta_{\mathrm{old}}}(x_{i,k+1}\mid x_{i,k})}.
\end{equation}
The clipped objective is
\begin{equation}
 \mathcal{L}_{\mathrm{GRPO}}=-\frac{1}{G|W|}
 \sum_{i=1}^{G}\sum_{k\in W}
 \min\!\left(\rho_{i,k}A_i,
 \operatorname{clip}(\rho_{i,k},1-\epsilon,1+\epsilon)A_i\right),
 \label{eq:grpo}
\end{equation}
with $\epsilon=10^{-4}$. We set the optional KL coefficient to zero in all reported runs, matching the experiment configuration.

\section{Experimental protocol}
\label{sec:experiments}

\subsection{Research questions}

The experiments address five questions: (RQ1) how the positions of sampling and optimization windows change the affect--semantic trade-off; (RQ2) whether semantic-first improves upon progressive scheduling and a non-RL affect-conditioned base; (RQ3) how it compares with full-path training in quality and cost; (RQ4) whether the conclusion persists at extreme affect targets and under human judgment; and (RQ5) how it transfers across window widths and a second backbone.

\subsection{Models, data, and training}

The primary backbone is FLUX.1-dev \cite{flux2024}; SD3.5-Medium provides the second-backbone evaluation \cite{sd35medium}. VA-EIT training uses the released EmotiCrafter prompt-pair corpus, which supplies neutral prompts, emotional rewrites, and their V--A coordinates \cite{emoticrafter2025}. We randomly split paired rows into approximately 98\% training and 2\% validation with seed 42 to train and select the adapter in Eq.~\eqref{eq:eit}. The same corpus supplies neutral prompts and V--A targets for GRPO. After adapter selection, we cache emotion-conditioned embeddings for all paired rows and use the complete cache across temporal schedules.

GRPO uses 25 sampling steps, 720$\times$720 output resolution, SDE noise scale $\eta=0.7$, timestep shift 3, 12 generations per prompt, and equal affect/semantic reward weights. The generation transformer is optimized with AdamW at learning rate $10^{-5}$, while the text encoders, VAE, and EIT remain frozen. Runs use eight NVIDIA A100 GPUs and training seeds 42, 123, and 3407. Each formal run comprises 10 additional warm-up iterations followed by 300 formal outer iterations. The separate 200-update pilot uses the same three training seeds. The progressive schedule advances its four-step window by one index every 14 iterations.

The 132 evaluation prompts do not overlap with the prompts used for VA-EIT or GRPO training. Evaluation crosses these neutral prompts with the 25-point grid $V,A\in\{-3,-1.5,0,1.5,3\}$, yielding 3,300 images per checkpoint. For each prompt, initial noise is shared across V--A targets and methods to control composition variation.

The baselines are: (1) the reproduced original EmotiCrafter configuration; (2) FLUX+VA-EIT without RL, which isolates the stronger backbone and adapter; (3) full-path Flow-GRPO; and (4) progressive MixGRPO. The proposed method uses the fixed late window $W(20)$ with the joint affect/semantic objective. All trained-checkpoint comparisons use full-trajectory ODE evaluation with trained weights at every step.

\subsection{Metrics}

V-Error and A-Error are mean absolute deviations between target values and predictions from the frozen CLIP-based affect regressors used in the reward pipeline. Independent affect evaluation uses frozen Qwen2.5-VL-7B-Instruct \cite{qwen25vl}, which receives the image and estimates V/A coordinates without the target, prompt, or method identity. Qwen V-MAE and A-MAE compare these image-only estimates with the target coordinates. CLIPScore is text--image cosine similarity reported after multiplication by 100 \cite{hessel2021clipscore}. TIFA uses visual question answering to measure prompt faithfulness and is reported on a 0--100 scale \cite{hu2023tifa}. For the formal comparison, a fixed question set is generated from each original prompt, filtered with UnifiedQA, and answered from each generated image by mPLUG-large; the same questions and answer keys are shared across methods, V--A targets, and repeats. Scores average question correctness within each image, then the 25 target variants and 132 prompts. CLIP-IQA measures no-reference perceptual quality on a 0--1 scale \cite{wang2023clipiqa}. DINO consistency is the mean pairwise cosine similarity, multiplied by 100, between DINOv2 embeddings of the 25 V--A variants generated for the same prompt \cite{oquab2024dinov2}; it measures whether content remains stable as affect changes. For a target-specific score, we average the similarity of that target image to the other 24 variants of the same prompt, then average over prompts. Table~\ref{tab:extreme} reports these scores at five selected targets, with means and sample SD across seeds; the main comparison averages over all 25 targets.

Semantic drift is the percentage of images with a replaced or absent principal subject. The expanded audit assesses prompt-specified identity, count, action, and spatial relations on shared applicable sets of 150, 130, 120, and 105 images per method, respectively. Both this audit and the five-target retention experiment in Section~\ref{sec:extreme} use two independent annotators and a third adjudicator to determine final binary correctness labels. The audit uses the same seed-42 images as the user study in Section~4.4. For each requirement, Cohen's $\kappa$ is computed by pooling the two annotators' independent, pre-adjudication labels across all six methods.

\subsection{Reporting protocol and user study}

Repeated RL results report means and sample SD over three training seeds; fixed-model sampling experiments use three sampling seeds. Automatic paired confidence intervals use 132 prompts as resampling units, retaining all 25 V--A variants and the fixed seed results within each prompt. We use 10,000 paired bootstrap resamples and percentile 95\% intervals. These intervals describe prompt-sampling uncertainty conditional on the evaluated runs. For the single EmotiCrafter checkpoint, the comparator remains fixed while the proposed method is averaged over its repeated runs.

The expanded user study includes 30 participants and a stratified random sample of 30 prompts from the 132-prompt test set, using four semantic strata: people, animals/natural scenes, man-made objects, and abstract/atmospheric content. Each prompt is evaluated at the four V--A corners and the neutral point, yielding 150 matched groups and 900 distinct seed-42 images across six methods. Each participant is randomly assigned 15 groups (90 images), and each group is evaluated in full by three different participants. The prompt and target are visible, method identities are hidden, and group order and within-group image order are randomized. The user study was exempt from ethics review, and all participants provided informed consent. Participants independently rate affect match, semantic preservation, and overall preference on a 1--5 scale.

For each participant, method, and rating dimension, we average scores over the 15 assigned groups. We report the mean and sample SD of these participant-level method means across 30 participants. For each baseline and rating dimension, we calculate 30 within-participant differences, defined as semantic-first minus the baseline, and construct two-sided 95\% paired-$t$ confidence intervals using 29 degrees of freedom.

\section{Results}
\label{sec:results}

\subsection{RQ1: Window position reveals a monotonic trade-off}

\begin{figure}[!htb]
\centering
\includegraphics[width=\textwidth]{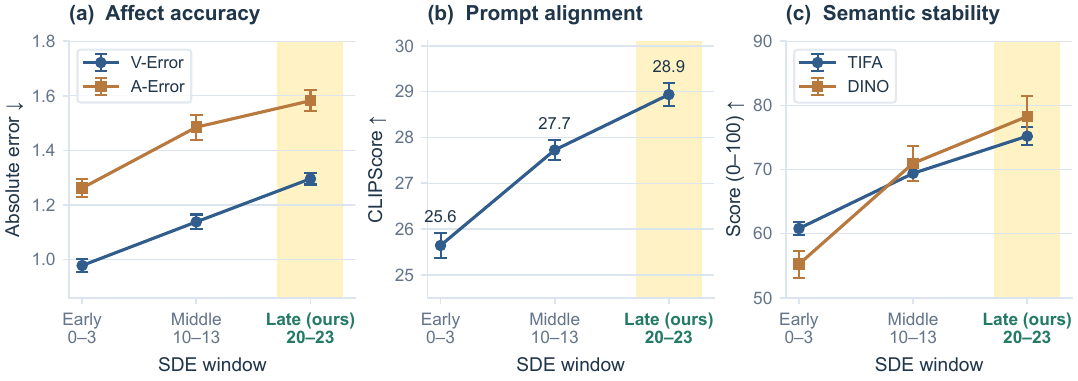}
\caption{Pilot trends as the SDE window moves from high-noise to low-noise steps. (a) V/A errors in the original $[-3,3]$ coordinate units; (b) CLIPScore, equal to cosine similarity multiplied by 100; (c) TIFA and DINO on their 0--100 presentation scales. Points show means and error bars show sample SD over training seeds 42, 123, and 3407, with 200 updates per run.}
\label{fig:pilot}
\end{figure}

Figure~\ref{fig:pilot} and Appendix Table~\ref{tab:pilot} report the 200-update pilot repeated with three training seeds. Moving the four-step SDE window from early to late raises mean V-Error from 0.978 to 1.296 and A-Error from 1.262 to 1.582, increases of 32.5\% and 25.4\%. At the same time, CLIPScore rises by 12.9\%, TIFA by 23.7\%, and DINO consistency by 41.7\%. The middle window lies between these endpoints on every reported metric mean. The three-seed pilot shows the affect--semantic trade-off across window positions, complemented by the separate exploration and optimization controls in Section~\ref{sec:controls}.

\subsection{RQ2--RQ3: Main comparison}

\begin{table}[!htb]
\centering
\begin{threeparttable}
\caption{Formal FLUX comparison, including TIFA. Values are means $\pm$ sample SD where repeated results are available.}
\label{tab:main}
\fontsize{9}{11}\selectfont
\setlength{\tabcolsep}{1.8pt}
\renewcommand{\arraystretch}{1.08}
\begin{tabular}{@{}lcccccc@{}}
\toprule
\rowcolor{tableblue}
Method & V-Error $\downarrow$ & A-Error $\downarrow$ & CLIP $\uparrow$ & DINO $\uparrow$ & IQA $\uparrow$ & TIFA $\uparrow$ \\
\midrule
EmotiCrafter & $1.510$ & $1.828$ & $27.74$ & $77.3$ & $0.917$ & $74.82$ \\
FLUX+VA-EIT & $1.402\mskip2mu\mathord{\pm}\mskip2mu0.048$ & $1.710\mskip2mu\mathord{\pm}\mskip2mu0.036$ & $29.27\mskip2mu\mathord{\pm}\mskip2mu0.27$ & $\mathbf{81.6}\mskip2mu\mathord{\pm}\mskip2mu0.9$ & $\mathbf{0.924}\mskip2mu\mathord{\pm}\mskip2mu0.024$ & $80.49\mskip2mu\mathord{\pm}\mskip2mu0.63$ \\
A: Full-path & $\mathbf{0.947}\mskip2mu\mathord{\pm}\mskip2mu0.046$ & $\mathbf{1.191}\mskip2mu\mathord{\pm}\mskip2mu0.037$ & $27.34\mskip2mu\mathord{\pm}\mskip2mu0.19$ & $64.7\mskip2mu\mathord{\pm}\mskip2mu6.2$ & $0.907\mskip2mu\mathord{\pm}\mskip2mu0.021$ & $79.75\mskip2mu\mathord{\pm}\mskip2mu1.08$ \\
B: Full-path & $0.963\mskip2mu\mathord{\pm}\mskip2mu0.041$ & $1.208\mskip2mu\mathord{\pm}\mskip2mu0.034$ & $27.58\mskip2mu\mathord{\pm}\mskip2mu0.21$ & $65.5\mskip2mu\mathord{\pm}\mskip2mu5.4$ & $0.906\mskip2mu\mathord{\pm}\mskip2mu0.019$ & $80.14\mskip2mu\mathord{\pm}\mskip2mu0.92$ \\
Progressive & $1.091\mskip2mu\mathord{\pm}\mskip2mu0.014$ & $1.386\mskip2mu\mathord{\pm}\mskip2mu0.014$ & $28.22\mskip2mu\mathord{\pm}\mskip2mu0.20$ & $74.4\mskip2mu\mathord{\pm}\mskip2mu0.7$ & $0.912\mskip2mu\mathord{\pm}\mskip2mu0.005$ & $78.53\mskip2mu\mathord{\pm}\mskip2mu0.71$ \\
Emotion-first & $0.974\mskip2mu\mathord{\pm}\mskip2mu0.028$ & $1.254\mskip2mu\mathord{\pm}\mskip2mu0.026$ & $26.41\mskip2mu\mathord{\pm}\mskip2mu0.16$ & $58.5\mskip2mu\mathord{\pm}\mskip2mu1.7$ & $0.893\mskip2mu\mathord{\pm}\mskip2mu0.014$ & $76.91\mskip2mu\mathord{\pm}\mskip2mu0.88$ \\
\rowcolor{oursrow}
Semantic-first & $1.052\mskip2mu\mathord{\pm}\mskip2mu0.044$ & $1.336\mskip2mu\mathord{\pm}\mskip2mu0.038$ & $\mathbf{29.51}\mskip2mu\mathord{\pm}\mskip2mu0.35$ & $79.6\mskip2mu\mathord{\pm}\mskip2mu2.1$ & $0.917\mskip2mu\mathord{\pm}\mskip2mu0.020$ & $\mathbf{82.34}\mskip2mu\mathord{\pm}\mskip2mu0.57$ \\
\bottomrule
\end{tabular}
\begin{tablenotes}[flushleft]\footnotesize
\item CLIPScore, DINO consistency and TIFA are multiplied by 100; CLIP-IQA is on a 0--1 scale. Bold marks the best mean in each column; yellow identifies our method F. FLUX+VA-EIT is the no-RL model. RL deviations are across three training seeds, and the no-RL deviation is across three sampling seeds. EmotiCrafter is a single fixed-checkpoint evaluation. The five regressor/perceptual metrics for EmotiCrafter and Progressive use their reference evaluation runs; A/B/E/F use the controlled-training runs, and the no-RL row uses S0. TIFA uses the formal evaluation images for all seven methods.
\end{tablenotes}
\end{threeparttable}
\end{table}

\begin{figure}[!htb]
\centering
\includegraphics[width=\linewidth]{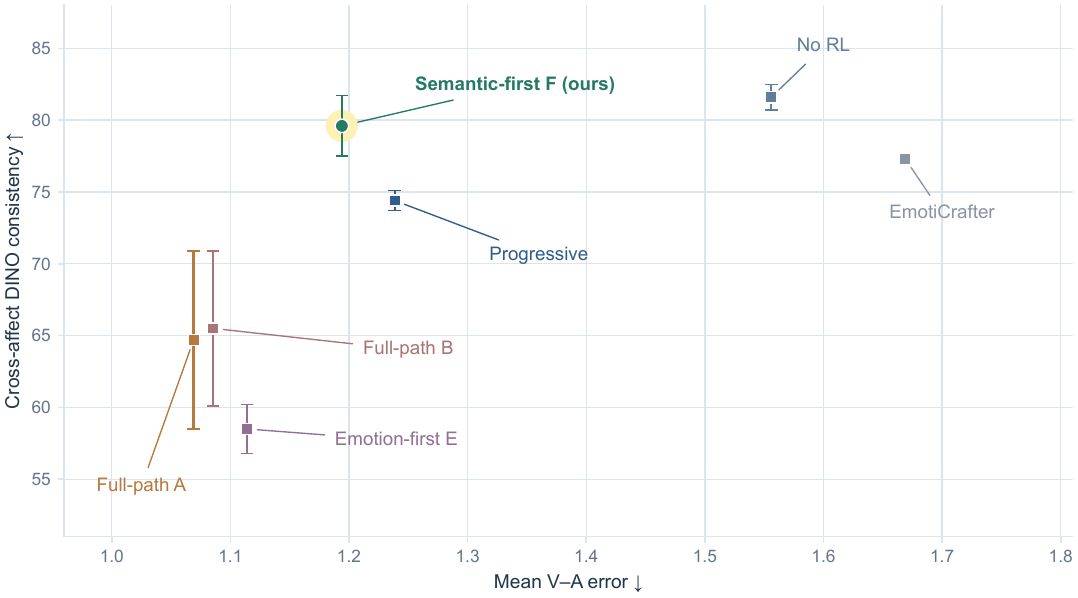}
\caption{Affect--semantic operating points on FLUX. The horizontal axis is the mean of V-Error and A-Error in V--A coordinate units; the vertical axis is pairwise DINO cosine similarity multiplied by 100. Lower horizontal values and higher vertical values are preferred. Vertical bars show the sample SD of DINO across the corresponding training or sampling repeats; EmotiCrafter is a single fixed-checkpoint result.}
\label{fig:pareto}
\end{figure}

Table~\ref{tab:main} gives the formal results, and Figure~\ref{fig:pareto} shows the affect--semantic operating points. Relative to the original full-path baseline A, semantic-first raises CLIPScore from 27.34 to 29.51 and DINO consistency from 64.7 to 79.6. Against the terminal-matched full-path B, the gains remain 1.93 CLIPScore points and 14.1 DINO points, with positive paired confidence intervals (Table~\ref{tab:control_ci}). Reward-regressor V/A errors are 1.052/1.336 for semantic-first and 0.963/1.208 for B. Independent evaluation in Section~\ref{sec:independent} assesses affect accuracy using a separate model.

Against progressive MixGRPO, semantic-first improves the reported affect, semantic, and image-quality means. Relative to the no-RL model, V-Error and A-Error decrease by 25.0\% and 21.9\%, while CLIPScore increases from 29.27 to 29.51. DINO consistency remains 79.6 and CLIP-IQA 0.917, compared with 81.6 and 0.924 before RL. The late window therefore adds substantial affect control while preserving high semantic consistency and image quality.

TIFA supplies a direct content-faithfulness check in the main evaluation. Semantic-first obtains the highest score, $82.34\pm0.57$, exceeding all six comparators with positive paired intervals (Table~\ref{tab:tifa_ci}). Gains are 1.85 points over no RL, 2.20 over matched full-path B, and 5.43 over emotion-first. These results extend the semantic advantage from feature similarity to question-based verification of prompt requirements.

\subsection{Separating exploration from policy optimization}
\label{sec:controls}

We separate the stochastic-transition set $S$ from the optimized set $U$, while matching the terminal transition (Table~\ref{tab:schedule_config}). The six configurations share the backbone, reward, training budget and evaluation procedure. Comparing D with C moves optimization from the first four to the last four nonterminal transitions while retaining full-path SDE exploration. CLIPScore rises by 1.04 and DINO by 10.7. Comparing F with D then confines exploration to those same late transitions, adding 1.39 CLIPScore points and 7.3 DINO points. Each of these semantic differences has a positive paired 95\% interval. Late optimization and late exploration thus make distinct contributions to content preservation.

\begin{table}[!htb]
\centering
\begin{threeparttable}
\caption{Controlled sampling and optimization configurations.}
\label{tab:schedule_config}
\small
\setlength{\tabcolsep}{3.2pt}
\renewcommand{\arraystretch}{1.08}
\begin{tabular}{lccc}
\toprule
\rowcolor{tableblue}
ID & SDE transitions & Optimized transitions & Transition 24 \\
\midrule
A & 0--24 & 0--23 & SDE \\
B & 0--23 & 0--23 & ODE \\
C & 0--23 & 0--3 & ODE \\
D & 0--23 & 20--23 & ODE \\
E & 0--3 & 0--3 & ODE \\
\rowcolor{oursrow}
F & 20--23 & 20--23 & ODE \\
\bottomrule
\end{tabular}
\begin{tablenotes}[flushleft]\footnotesize
\item A is the original full-path baseline; B matches the terminal ODE transition. C/D retain full-path exploration with early/late optimization. E/F restrict both sampling and optimization to the early/late window. All trajectories have 25 transitions.
\end{tablenotes}
\end{threeparttable}
\end{table}

\begin{table}[!htb]
\centering
\begin{threeparttable}
\caption{Effects of sampling and optimization windows on FLUX. Means $\pm$ sample SD over three training seeds.}
\label{tab:controlled}
\footnotesize
\setlength{\tabcolsep}{3.2pt}
\renewcommand{\arraystretch}{1.08}
\begin{tabular}{lccccc}
\toprule
\rowcolor{tableblue}
ID & V-Error $\downarrow$ & A-Error $\downarrow$ & CLIPScore $\uparrow$ & DINO $\uparrow$ & CLIP-IQA $\uparrow$ \\
\midrule
A & $0.947\pm0.046$ & $1.191\pm0.037$ & $27.34\pm0.19$ & $64.7\pm6.2$ & $0.907\pm0.021$ \\
B & $0.963\pm0.041$ & $1.208\pm0.034$ & $27.58\pm0.21$ & $65.5\pm5.4$ & $0.906\pm0.019$ \\
C & $0.992\pm0.058$ & $1.251\pm0.047$ & $27.08\pm0.24$ & $61.6\pm3.1$ & $0.899\pm0.017$ \\
D & $1.024\pm0.052$ & $1.294\pm0.043$ & $28.12\pm0.23$ & $72.3\pm2.4$ & $0.913\pm0.016$ \\
E & $0.974\pm0.028$ & $1.254\pm0.026$ & $26.41\pm0.16$ & $58.5\pm1.7$ & $0.893\pm0.014$ \\
\rowcolor{oursrow}
F & $1.052\pm0.044$ & $1.336\pm0.038$ & $29.51\pm0.35$ & $79.6\pm2.1$ & $0.917\pm0.020$ \\
\bottomrule
\end{tabular}
\begin{tablenotes}[flushleft]\footnotesize
\item Configurations are defined in Table~\ref{tab:schedule_config}. All checkpoints use the same full-trajectory ODE evaluation.
\end{tablenotes}
\end{threeparttable}
\end{table}

The fixed-model intervention isolates sampling behavior without policy updates (Table~\ref{tab:frozen}). Moving stochastic sampling from steps 0--3 (S1) to 20--23 (S2) raises CLIPScore by 1.17 and DINO by 7.6, with positive paired intervals. Together with the trained controls, this shows how the temporal location of exploration affects semantics both before and after optimization. Matching the terminal step alone (B versus A) changes CLIPScore by 0.24 and DINO by 0.8; the F-versus-B comparison retains the larger late-window gains.

Training-window selection and inference-weight usage are also separated. F-all applies trained weights throughout ODE inference; F-window applies the same weights only at steps 20--23. F-all has better means on all five original metrics, including CLIPScore 29.51 versus 29.24 and DINO 79.6 versus 79.0 (Table~\ref{tab:weights}). The method therefore uses selective policy-loss timesteps during training and a shared trained policy throughout inference.

\subsection{Independent affect evaluation}
\label{sec:independent}

The frozen image-only Qwen evaluator gives semantic-first the lowest mean V-MAE and A-MAE, 1.284 and 1.479 (Table~\ref{tab:qwen}). Relative to no RL, these errors fall by 12.2\% and 7.7\%. Both error differences have negative paired intervals against EmotiCrafter, no RL, both full-path baselines, and progressive scheduling. Against emotion-first, the V-MAE difference is $-0.045$ with a 95\% interval of $[-0.083,-0.009]$; A-MAE is lower by 0.027 in the means (Table~\ref{tab:qwen_ci}). These results establish affect gains under the independent Qwen evaluator alongside the prompt-faithfulness and semantic-consistency improvements.

\begin{table}[!htb]
\centering
\begin{threeparttable}
\caption{Independent image-only affect evaluation with Qwen2.5-VL-7B-Instruct.}
\label{tab:qwen}
\small
\setlength{\tabcolsep}{3.2pt}
\renewcommand{\arraystretch}{1.08}
\begin{tabular}{lcc}
\toprule
\rowcolor{tableblue}
Method & Qwen V-MAE $\downarrow$ & Qwen A-MAE $\downarrow$ \\
\midrule
EmotiCrafter & $1.541$ & $1.685$ \\
FLUX+VA-EIT (no RL) & $1.463\pm0.054$ & $1.602\pm0.071$ \\
Flow-GRPO A & $1.412\pm0.082$ & $1.637\pm0.096$ \\
Full-path B & $1.395\pm0.059$ & $1.571\pm0.043$ \\
Progressive & $1.367\pm0.047$ & $1.548\pm0.068$ \\
Emotion-first E & $1.329\pm0.062$ & $1.506\pm0.051$ \\
\rowcolor{oursrow}
Semantic-first F & $\mathbf{1.284\pm0.036}$ & $\mathbf{1.479\pm0.039}$ \\
\bottomrule
\end{tabular}
\begin{tablenotes}[flushleft]\footnotesize
\item Errors are measured in the original V--A units, with targets in $[-3,3]^2$. Qwen is frozen and receives only the image. EmotiCrafter is evaluated once; other rows report mean $\pm$ sample SD across the existing three model/sampling repeats.
\end{tablenotes}
\end{threeparttable}
\end{table}

\subsection{Order, drift, and the meaning of ``semantic-first''}

Semantic-first achieves the highest accuracy on all four audit dimensions (Table~\ref{tab:order}). It preserves subject identity in 146 of 150 images, giving 97.3\% accuracy and 2.7\% drift, compared with drift rates of 30.0\% for Emotion-first and 26.7\% for full-path Flow-GRPO. Count, action, and spatial-relation accuracies are 97.7\%, 95.8\%, and 95.2\%, respectively. The semantic advantage therefore covers prompt-specified entities and their composition, beyond subject presence alone.

\begin{table}[!htb]
\centering
\begin{threeparttable}
\caption{Expanded semantic audit: accuracy by prompt requirement and subject-drift rate (\%).}
\label{tab:order}
\footnotesize
\setlength{\tabcolsep}{3.2pt}
\renewcommand{\arraystretch}{1.08}
\begin{tabular}{lccccc}
\toprule
\rowcolor{tableblue}
Method & Identity & Count & Action & Spatial & Drift $\downarrow$ \\
\midrule
EmotiCrafter & 72.0 & 67.7 & 65.8 & 65.7 & 28.0 \\
FLUX+VA-EIT (no RL) & 83.3 & 83.8 & 81.7 & 81.0 & 16.7 \\
Flow-GRPO & 73.3 & 70.8 & 70.8 & 70.5 & 26.7 \\
Progressive & 80.0 & 79.2 & 79.2 & 76.2 & 20.0 \\
Emotion-first & 70.0 & 66.9 & 63.3 & 63.8 & 30.0 \\
\rowcolor{oursrow}
Semantic-first (ours) & 97.3 & 97.7 & 95.8 & 95.2 & 2.7 \\
\bottomrule
\end{tabular}
\begin{tablenotes}[flushleft]\footnotesize
\item The same 150 images per method are audited. Applicable denominators are 150, 130, 120 and 105 for identity, count, action and spatial relations, respectively, and are identical across methods. Correct counts for our method are 146, 127, 115 and 100. Identity drift is $100\%-\mathrm{identity\ accuracy}$. Cohen\textquotesingle s $\kappa$ on the two annotators\textquotesingle{} independent correctness labels is 0.78, 0.72, 0.67 and 0.64, respectively; a third annotator resolves disagreements.
\end{tablenotes}
\end{threeparttable}
\end{table}

\subsection{RQ4: Extreme V--A targets}
\label{sec:extreme}

\begin{figure}[!htb]
\centering
\includegraphics[width=0.93\textwidth]{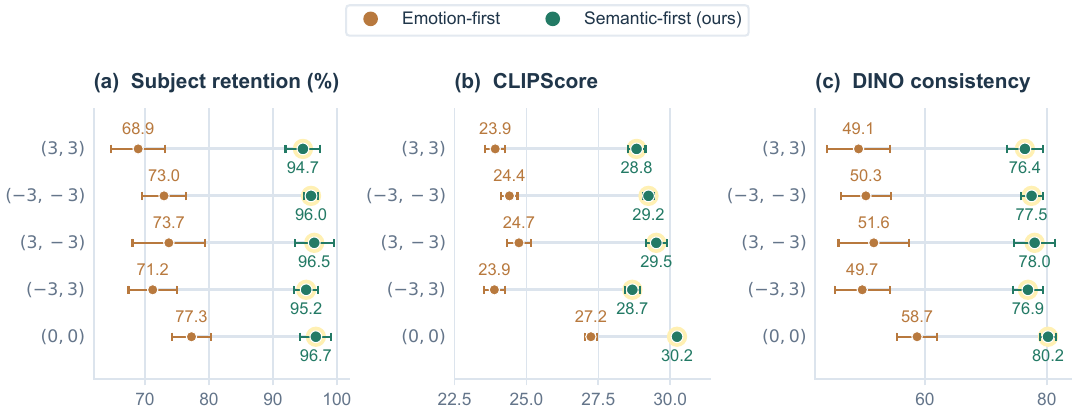}
\caption{Five-target evaluation at the four corners and center of the V--A grid. Panels report (a) principal-subject retention in percent, (b) CLIPScore (cosine similarity $\times100$), and (c) DINO consistency (cosine similarity $\times100$). Markers show means and horizontal error bars show sample SD across three seeds. Each method and target includes 132 images per seed (396 total).}
\label{fig:extreme}
\end{figure}

Figure~\ref{fig:extreme} and Appendix Table~\ref{tab:extreme} compare the two temporal orders at the four V--A corners and the neutral point. Averaged over these five targets, semantic-first raises CLIPScore from 24.838 to 29.303 (+18.0\%), DINO consistency from 51.88 to 77.80 (+50.0\%), and subject retention from 72.83\% to 95.81\% (+22.98 percentage points). V-Error increases by 10.6\%, while A-Error changes by only 0.9\%. Late stochasticity preserves the principal subject at all five targets, including the extreme corners of the control space.

\begin{figure}[!tp]
\centering
\includegraphics[width=\textwidth]{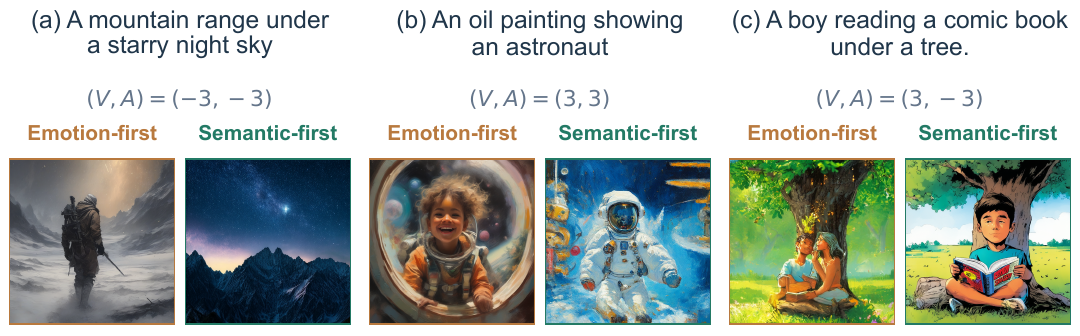}
\caption{Paired scheduling comparison at three V--A targets. Each pair shares the prompt, target, initial noise, and sampling parameters. Emotion-first changes subject or composition; Semantic-first preserves the mountain landscape, astronaut, and boy--book--tree relations.}
\label{fig:qualitative}
\vspace{3pt}
\includegraphics[width=\textwidth]{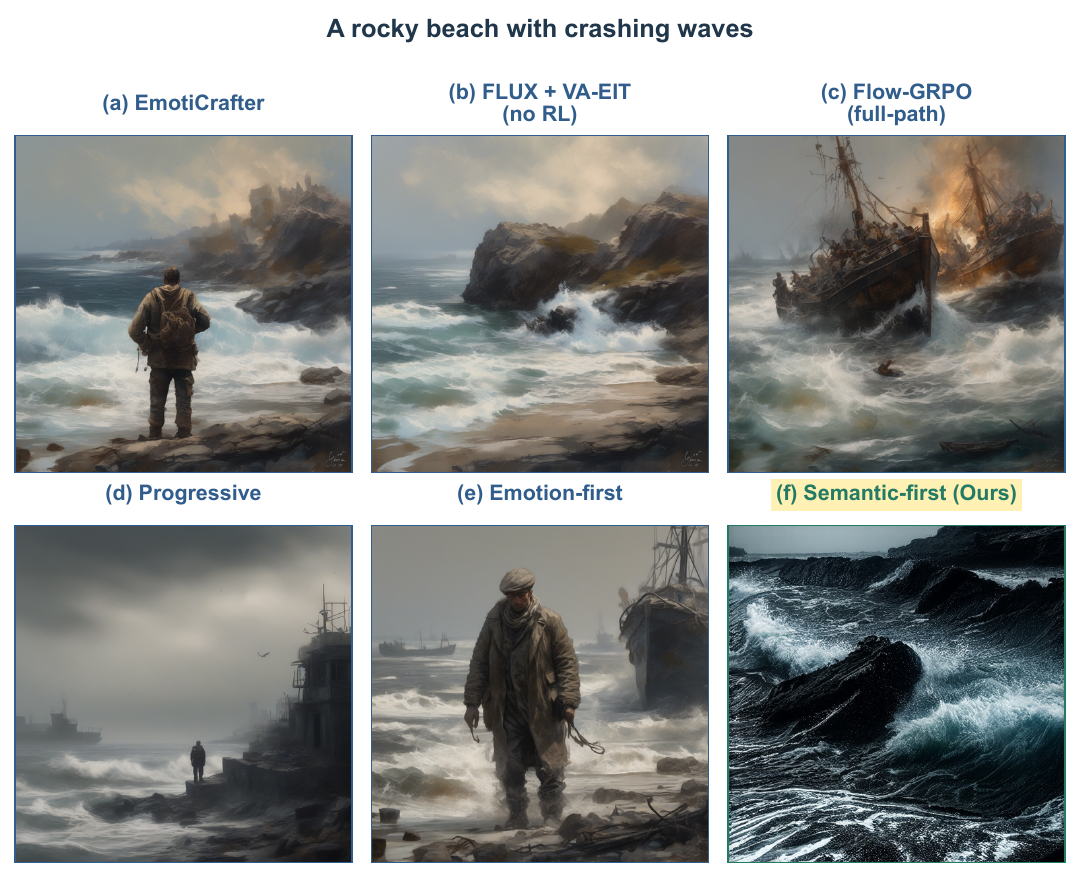}
\caption{Six-method comparison for ``A rocky beach with crashing waves,'' with $(V,A)=(-3,-3)$ and identical initial noise tensors. Each panel displays the complete output image at the same size.}
\label{fig:qualitative_baselines}
\end{figure}

Figure~\ref{fig:qualitative} shows three matched examples of semantic drift. At $(-3,-3)$, Emotion-first inserts a foreground traveler and suppresses most of the starry sky; Semantic-first retains the mountain range and night-sky composition. At $(3,3)$, the shift is milder: Emotion-first produces a close-up, child-centered portrait with weaker oil-painting appearance, while Semantic-first retains an unambiguous astronaut and painterly rendering. At $(3,-3)$, Emotion-first replaces one boy reading under a tree with two people and an unclear reading action; Semantic-first preserves the boy, book, tree, and comic-book style. These cases illustrate how early stochasticity can change subject count, identity, action, or composition, while late stochasticity more often preserves these semantic anchors during appearance changes, consistent with the drift and retention results in Tables~\ref{tab:order} and~\ref{tab:extreme}.

Figure~\ref{fig:qualitative_baselines} compares six methods for the prompt ``A rocky beach with crashing waves.'' Both the no-RL baseline and Semantic-first retain the rocky coast and breaking waves as the main content. Semantic-first renders a darker sea with pronounced wave crests and foam. EmotiCrafter and Emotion-first introduce foreground people; Progressive adds a person and waterfront structures, while full-path Flow-GRPO shifts the composition toward ships and fire. This example makes changes in scene content and appearance directly visible alongside the semantic audit, TIFA, and human-study results.

\subsection{RQ5(a): Window-width sensitivity}
\label{sec:width}

We vary late-window width over $w\in\{2,4,6,8\}$ while fixing the final stochastic/optimized transition at 23: $W_w=\{24-w,\ldots,23\}$. This gives windows 22--23, 20--23, 18--23, and 16--23, with an ODE terminal transition in every case. All runs share the F configuration and training budget; W4 reuses F-all. Table~\ref{tab:width} reports the original metrics, and Appendix Table~\ref{tab:width_qwen} reports independent affect evaluation.

W2 obtains the highest CLIPScore, DINO, and CLIP-IQA means, while W6 obtains the lowest affect errors. Relative to W4, W2 increases CLIPScore by 0.33 and DINO by 1.3 but increases original V/A errors by 0.045/0.035. W6 reduces those errors by 0.039/0.047 while reducing CLIPScore by 0.88 and DINO by 1.8. These differences have intervals excluding zero (Table~\ref{tab:width_ci}). Independent Qwen A-MAE also improves for W6 by 0.038 with interval $[-0.069,-0.006]$. W8 has worse means than W4 on all seven metrics. The results characterize width as a controllable affect--semantic trade-off and support W4 as the default balance between these objectives.

Training time increases from 13.85 seconds per iteration at W2 to 29.68 at W8 (Table~\ref{tab:width_cost}). W6 and W8 cost 29.6\% and 58.2\% more than W4, respectively. Width therefore controls both computational cost and the affect--semantic operating point.

\begin{table}[!htb]
\centering
\begin{threeparttable}
\caption{Late-window width sensitivity on FLUX.}
\label{tab:width}
\footnotesize
\setlength{\tabcolsep}{2.7pt}
\renewcommand{\arraystretch}{1.08}
\begin{tabular}{lccccc}
\toprule
\rowcolor{tableblue}
Width (steps) & V-Error $\downarrow$ & A-Error $\downarrow$ & CLIP $\uparrow$ & DINO $\uparrow$ & IQA $\uparrow$ \\
\midrule
W2 (22--23) & $1.097\pm0.062$ & $1.371\pm0.051$ & $\mathbf{29.84\pm0.47}$ & $\mathbf{80.9\pm2.6}$ & $\mathbf{0.921\pm0.013}$ \\
\rowcolor{oursrow}
W4 (20--23) & $1.052\pm0.044$ & $1.336\pm0.038$ & $29.51\pm0.35$ & $79.6\pm2.1$ & $0.917\pm0.020$ \\
W6 (18--23) & $\mathbf{1.013\pm0.059}$ & $\mathbf{1.289\pm0.047}$ & $28.63\pm0.51$ & $77.8\pm2.4$ & $0.910\pm0.018$ \\
W8 (16--23) & $1.124\pm0.073$ & $1.409\pm0.066$ & $27.91\pm0.68$ & $73.4\pm3.2$ & $0.898\pm0.031$ \\
\bottomrule
\end{tabular}
\begin{tablenotes}[flushleft]\footnotesize
\item The last stochastic/optimized transition is fixed at 23; transition 24 is ODE. Means $\pm$ sample SD are over three training seeds. W4 repeats F-all.
\end{tablenotes}
\end{threeparttable}
\end{table}

\subsection{RQ5(b): Replication on SD3.5-Medium}
\label{sec:backbone}

SD3.5-Medium uses a shared pretrained T5-branch VA-EIT to compare no RL (M0), terminal-matched full-path training (M1), early-window training (M2), and semantic-first (M3). Three-seed RL runs optimize transitions 0--23 for M1 and four-transition windows for M2/M3. All configurations use 25 steps, 720$\times$720 resolution, CFG 4.5, an ODE terminal transition, and full-trajectory ODE evaluation, sharing the FLUX prompts, V--A grid, rewards, and evaluators.

Semantic-first achieves the best mean on all seven metrics among the RL configurations (Tables~\ref{tab:sd3} and~\ref{tab:sd3_qwen}). Against matched full-path M1, it improves CLIPScore by 1.29 and DINO by 4.2, while reducing original V/A errors by 0.062/0.157 and Qwen errors by 0.103/0.092; all six paired intervals support these improvements (Table~\ref{tab:sd3_ci}). Against early-window M2, CLIPScore and DINO rise by 2.38 and 7.6, and both original and independent valence errors improve with negative intervals. Relative to no RL, original V/A errors decrease by 16.2\%/21.6\%, while DINO remains 79.4 and CLIP-IQA 0.920. These results reproduce the semantic-first scheduling advantage on a second flow-matching backbone.

\begin{table}[!htb]
\centering
\begin{threeparttable}
\caption{Second-backbone evaluation on SD3.5-Medium.}
\label{tab:sd3}
\footnotesize
\setlength{\tabcolsep}{1.9pt}
\renewcommand{\arraystretch}{1.08}
\begin{tabular}{lccccc}
\toprule
\rowcolor{tableblue}
Method & V-Error $\downarrow$ & A-Error $\downarrow$ & CLIP $\uparrow$ & DINO $\uparrow$ & IQA $\uparrow$ \\
\midrule
M0: no RL & $1.274\pm0.058$ & $1.651\pm0.043$ & $29.07\pm0.16$ & $\mathbf{81.9\pm0.6}$ & $\mathbf{0.925\pm0.007}$ \\
M1: Full-path & $1.129\pm0.091$ & $1.452\pm0.077$ & $28.42\pm0.25$ & $75.2\pm2.1$ & $0.909\pm0.019$ \\
M2: Early & $1.092\pm0.047$ & $1.318\pm0.039$ & $27.33\pm0.19$ & $71.8\pm1.4$ & $0.916\pm0.012$ \\
\rowcolor{oursrow}
M3: Late (ours) & $\mathbf{1.067\pm0.031}$ & $\mathbf{1.295\pm0.028}$ & $\mathbf{29.71\pm0.21}$ & $79.4\pm1.2$ & $0.920\pm0.009$ \\
\bottomrule
\end{tabular}
\begin{tablenotes}[flushleft]\footnotesize
\item M1 uses SDE sampling/optimization at 0--23; M2 at 0--3; M3 at 20--23. Transition 24 is ODE for every configuration. M1--M3 report sample SD across three training seeds; M0 reports sample SD across three sampling seeds.
\end{tablenotes}
\end{threeparttable}
\end{table}

\subsection{Efficiency}

\begin{figure}[!htb]
\centering
\includegraphics[width=\textwidth]{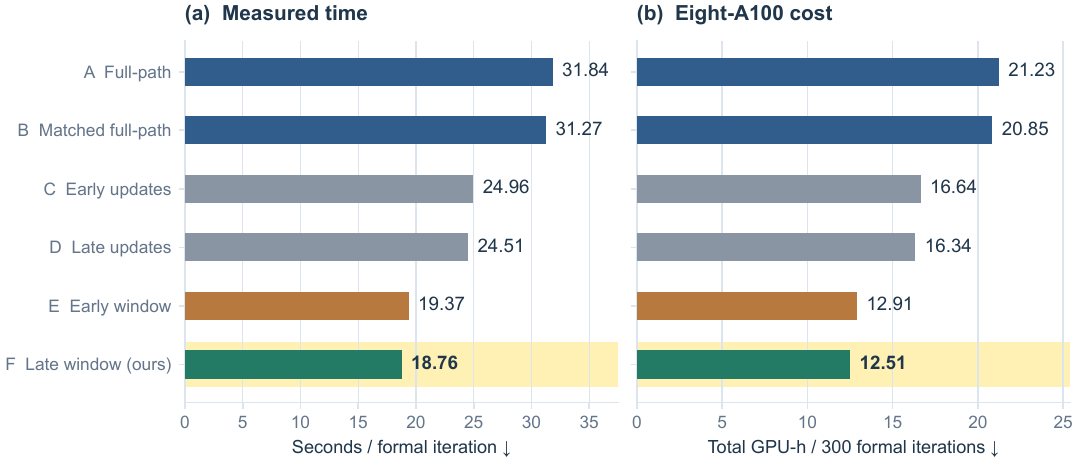}
\caption{Measured training time and total GPU cost on eight A100 GPUs. All configurations use 300 formal iterations after 10 additional warm-up iterations. A/B use 24 policy-loss transitions per trajectory; C--F use four. Yellow identifies F.}
\label{fig:efficiency}
\end{figure}

Four-step schedules reduce train-forward NFE per rank and outer iteration from $12\times24=288$ to $12\times4=48$, an 83.3\% reduction in that component. End-to-end cost also includes rollout, deterministic ODE evaluations, decoding, reward inference, and communication. On eight A100 GPUs (Table~\ref{tab:efficiency}, Figure~\ref{fig:efficiency}), Semantic-first reduces formal iteration time from 31.84 to 18.76 seconds relative to A, a 41.1\% reduction and $1.70\times$ speedup. Relative to terminal-matched B, the reduction is 40.0\% ($1.67\times$). Its 300 formal iterations require 1.56 wall-clock hours and 12.51 total GPU-hours.

The controls separate the sources of the saving. B to D retains full-path stochastic exploration and reduces optimization from 24 to four transitions, lowering time by 21.6\%. D to F then restricts exploration to the late window and lowers time by a further 23.5\%. These consecutive reductions compose to the 40.0\% B-to-F saving. Timing covers rollout, decoding, reward inference, policy updates and distributed execution during the 300 formal iterations; the 10 additional warm-up iterations precede the timed interval.

\subsection{User study}

\begin{figure}[!htb]
\centering
\includegraphics[width=\textwidth]{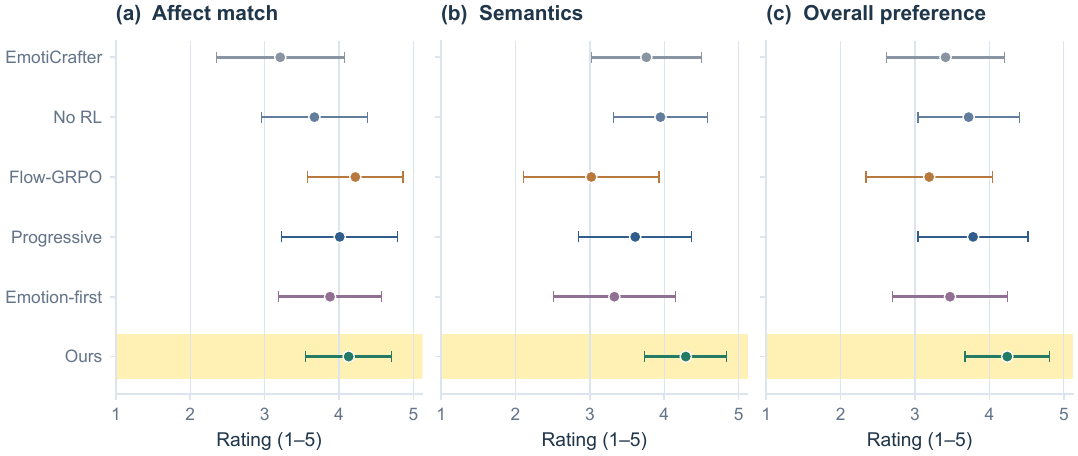}
\caption{Expanded human evaluation on a 1--5 scale. Thirty participants rate 900 images, with three ratings per image. Points show the mean of participant-level method means; error bars show their sample SD. Yellow identifies Semantic-first.}
\label{fig:user}
\end{figure}

The expanded study favors semantic-first on semantic preservation and overall preference (Table~\ref{tab:user}, Figure~\ref{fig:user}). Its mean ratings are 4.29 and 4.24, respectively, and all paired differences against the five comparators have positive confidence intervals (Table~\ref{tab:human_ci}). Relative to Flow-GRPO, the gains are 1.27 points for semantic preservation and 1.05 for overall preference. Affect match remains high at 4.13, compared with 4.22 for Flow-GRPO; its paired difference is $-0.09$ with interval $[-0.31,0.13]$. The user study therefore supports the practical value of semantic-first generation: strong affect matching together with better preservation of requested content and higher overall preference.

\begin{table}[!htb]
\centering
\begin{threeparttable}
\caption{Expanded human evaluation: 30 participants, 900 images, and three ratings per image.}
\label{tab:user}
\footnotesize
\setlength{\tabcolsep}{3.2pt}
\renewcommand{\arraystretch}{1.08}
\begin{tabular}{lccc}
\toprule
\rowcolor{tableblue}
Method & Affect match & Semantic preservation & Overall preference \\
\midrule
EmotiCrafter & $3.21\pm0.86$ & $3.76\pm0.74$ & $3.41\pm0.79$ \\
FLUX+VA-EIT (no RL) & $3.67\pm0.71$ & $3.95\pm0.63$ & $3.72\pm0.68$ \\
Flow-GRPO & $\mathbf{4.22\pm0.64}$ & $3.02\pm0.91$ & $3.19\pm0.85$ \\
Progressive & $4.01\pm0.78$ & $3.61\pm0.76$ & $3.78\pm0.74$ \\
Emotion-first & $3.88\pm0.69$ & $3.33\pm0.82$ & $3.47\pm0.77$ \\
\rowcolor{oursrow}
Semantic-first (ours) & $4.13\pm0.58$ & $\mathbf{4.29\pm0.55}$ & $\mathbf{4.24\pm0.57}$ \\
\bottomrule
\end{tabular}
\begin{tablenotes}[flushleft]\footnotesize
\item Scores use a 1--5 scale. Entries are means $\pm$ sample SD of participant-level method means. Each participant evaluates 15 matched six-method groups (90 images).
\end{tablenotes}
\end{threeparttable}
\end{table}

\section{Engineering discussion}
\label{sec:discussion}

The controls identify denoising time as an attribute-control resource under the shared joint reward in Eq.~\eqref{eq:advantage}. With full-path exploration fixed, late optimization improves semantic consistency; restricting exploration to the same late window adds a further gain. Moving stochasticity out of the high-noise phase also preserves content before policy updates. Temporal allocation thus protects prompt faithfulness, subject retention, and cross-affect consistency through both exploration and learning. This agrees with MixGRPO's time-dependent optimization signals \cite{mixgrpo2025} and extends the scheduling result from FLUX to SD3.5-Medium.

Affect rankings depend on the evaluator. The reward regressors favor full-path training, whereas independent image-only Qwen evaluation favors semantic-first. A plausible explanation is the difference in optimization exposure: the regressors also supply the training reward, which full-path training optimizes across more transitions, while Qwen provides a separate affect estimate without participating in training. This can favor full-path performance under the optimized evaluator without producing the same ordering under Qwen. Human affect-match ratings are 4.22 for full-path Flow-GRPO and 4.13 for semantic-first; the paired difference has a 95\% interval of $[-0.31,0.13]$, so this study does not establish a difference in affect match between them. Semantic-first improves semantic preservation and overall preference in the human study, with positive paired intervals against every comparator. Together with TIFA and the semantic audit, these results show a consistent content-preservation advantage and independently evaluated affect gains.

The width study makes the operating choice explicit. A two-step late window prioritizes semantic consistency and low cost; a six-step window strengthens affect control with a larger semantic and computational cost. The four-step default balances these objectives. Equation~\eqref{eq:pareto} provides the schedule-selection criterion for applications with different content-preservation and computational requirements.

\section{Limitations}
\label{sec:limitations}

The study evaluates FLUX.1-dev and SD3.5-Medium with continuous valence--arousal conditioning and 25-step sampling. The default window 20--23 is tied to this sampling budget and noise schedule; changing either requires reassessing window placement. Window comparisons measure scheduling effects under the inherited time-dependent SDE noise schedule, including the noise magnitude associated with each window. Widths from two to eight transitions characterize the tested operating range. The experiments establish scheduling effects on the affect--semantic trade-off, but do not establish an independent factorization of semantic and affective variables along denoising time. Future work can calibrate both affect evaluators against human V--A annotations to test the source of their ranking differences.

\section{Conclusion}
\label{sec:conclusion}

We studied the temporal allocation of semantic formation and affective exploration in flow-matching image generation. During RL training, the semantic-first schedule uses ODE sampling at high noise followed by SDE exploration and GRPO optimization at low noise. Separate sampling and optimization controls establish the contribution of each temporal choice, and width tests characterize its affect--semantic and computational trade-offs. On FLUX, semantic-first achieves the highest main-comparison TIFA score, improves independent Qwen affect evaluation, and obtains the highest semantic-preservation and overall-preference ratings. Its expanded audit records 2.7\% subject drift, while GRPO training time is reduced by 41.1\% relative to original full-path Flow-GRPO. Replication on SD3.5-Medium confirms the scheduling advantage on a second backbone. Probability-flow time is therefore a practical control dimension for reliable emotional image generation.

\section*{Declaration of competing interest}
The authors declare that they have no known competing financial interests or personal relationships that could have appeared to influence the work reported in this paper.

\section*{Data and code availability}
The implementation, experiment configurations, plotting code, complete 132-prompt evaluation list, and training-data split information are available in our \href{https://anonymous.4open.science/r/TemporalAffect-MixGRPO-4347/}{anonymous project repository}. It also provides the full Qwen evaluation prompt, output-parsing and failure-handling rules, and fixed TIFA question--answer sets. The manuscript source package includes the aggregate results used in the tables and figures. Model weights and third-party datasets remain subject to their original licenses.

\FloatBarrier
\appendix
\setcounter{table}{0}
\renewcommand{\thetable}{A.\arabic{table}}
\renewcommand{\theHtable}{appendix.A.\arabic{table}}

\section{Paired comparisons and additional controls}
\label{app:paired}
\begin{table}[H]
\centering
\begin{threeparttable}
\caption{Paired FLUX differences and 95\% prompt-bootstrap confidence intervals.}
\label{tab:control_ci}
\scriptsize
\setlength{\tabcolsep}{2.8pt}
\renewcommand{\arraystretch}{1.08}
\begin{tabular}{lcccc}
\toprule
\rowcolor{tableblue}
Comparison & CLIPScore & DINO & V-Error & A-Error \\
\midrule
F$-$B & $+1.93\;[1.68,\,2.18]$ & $+14.1\;[12.4,\,15.8]$ & $+0.089\;[0.048,\,0.130]$ & $+0.128\;[0.085,\,0.171]$ \\
F$-$E & $+3.10\;[2.78,\,3.42]$ & $+21.1\;[19.7,\,22.5]$ & $+0.078\;[0.043,\,0.113]$ & $+0.082\;[0.048,\,0.116]$ \\
D$-$C & $+1.04\;[0.76,\,1.32]$ & $+10.7\;[9.4,\,12.0]$ & $+0.032\;[0.012,\,0.052]$ & $+0.043\;[0.022,\,0.064]$ \\
F$-$D & $+1.39\;[1.09,\,1.69]$ & $+7.3\;[6.2,\,8.4]$ & $+0.028\;[0.011,\,0.045]$ & $+0.042\;[0.021,\,0.063]$ \\
S2$-$S1 & $+1.17\;[0.91,\,1.43]$ & $+7.6\;[6.4,\,8.8]$ & $-0.028\;[-0.058,\,0.002]$ & $-0.045\;[-0.072,\,-0.018]$ \\
B$-$A & $+0.24\;[0.03,\,0.45]$ & $+0.8\;[-0.3,\,1.9]$ & $+0.016\;[-0.008,\,0.040]$ & $+0.017\;[-0.007,\,0.041]$ \\
E$-$C & $-0.67\;[-0.89,\,-0.45]$ & $-3.1\;[-4.2,\,-2.0]$ & $-0.018\;[-0.038,\,0.002]$ & $+0.003\;[-0.019,\,0.025]$ \\
\bottomrule
\end{tabular}
\begin{tablenotes}[flushleft]\footnotesize
\item Each entry is the first configuration minus the second. The 132 prompts are the resampling units; all 25 V--A targets and the fixed seed results remain grouped. Intervals quantify prompt-sampling uncertainty conditional on these model runs.
\end{tablenotes}
\end{threeparttable}
\end{table}

\begin{table}[H]
\centering
\begin{threeparttable}
\caption{Independent affect evaluation: Ours minus each baseline, with 95\% prompt-bootstrap confidence intervals.}
\label{tab:qwen_ci}
\footnotesize
\setlength{\tabcolsep}{3.2pt}
\renewcommand{\arraystretch}{1.08}
\begin{tabular}{lcc}
\toprule
\rowcolor{tableblue}
Baseline & Qwen V-MAE difference & Qwen A-MAE difference \\
\midrule
EmotiCrafter & $-0.257\;[-0.312,\,-0.191]$ & $-0.206\;[-0.274,\,-0.141]$ \\
FLUX+VA-EIT (no RL) & $-0.179\;[-0.221,\,-0.132]$ & $-0.123\;[-0.164,\,-0.087]$ \\
Flow-GRPO A & $-0.128\;[-0.171,\,-0.082]$ & $-0.158\;[-0.219,\,-0.102]$ \\
Full-path B & $-0.111\;[-0.149,\,-0.077]$ & $-0.092\;[-0.134,\,-0.055]$ \\
Progressive & $-0.083\;[-0.124,\,-0.041]$ & $-0.069\;[-0.116,\,-0.018]$ \\
Emotion-first E & $-0.045\;[-0.083,\,-0.009]$ & $-0.027\;[-0.068,\,0.019]$ \\
\bottomrule
\end{tabular}
\begin{tablenotes}[flushleft]\footnotesize
\item Negative differences favor Semantic-first. Paired resampling uses 132 prompts with their 25 targets and fixed model repeats grouped. EmotiCrafter remains a single fixed-checkpoint comparator.
\end{tablenotes}
\end{threeparttable}
\end{table}

\begin{table}[H]
\centering
\begin{threeparttable}
\caption{TIFA paired differences: Semantic-first minus each baseline.}
\label{tab:tifa_ci}
\small
\setlength{\tabcolsep}{3.2pt}
\renewcommand{\arraystretch}{1.08}
\begin{tabular}{lcc}
\toprule
\rowcolor{tableblue}
Baseline & Difference & 95\% CI \\
\midrule
EmotiCrafter & $+7.52$ & $[7.18,\,7.89]$ \\
FLUX+VA-EIT (no RL) & $+1.85$ & $[1.49,\,2.18]$ \\
Flow-GRPO A & $+2.59$ & $[2.21,\,2.93]$ \\
Full-path B & $+2.20$ & $[1.86,\,2.57]$ \\
Progressive & $+3.81$ & $[3.47,\,4.16]$ \\
Emotion-first E & $+5.43$ & $[5.09,\,5.81]$ \\
\bottomrule
\end{tabular}
\begin{tablenotes}[flushleft]\footnotesize
\item Differences are TIFA score points on the 0--100 presentation scale. All six intervals are positive. The prompt-bootstrap protocol is shared with the other automatic paired comparisons.
\end{tablenotes}
\end{threeparttable}
\end{table}

\begin{table}[H]
\centering
\begin{threeparttable}
\caption{Human evaluation: Ours minus each baseline, with paired 95\% confidence intervals.}
\label{tab:human_ci}
\scriptsize
\setlength{\tabcolsep}{3.2pt}
\renewcommand{\arraystretch}{1.08}
\begin{tabular}{lccc}
\toprule
\rowcolor{tableblue}
Baseline & Affect match & Semantic preservation & Overall preference \\
\midrule
EmotiCrafter & $+0.92\;[0.61,\,1.23]$ & $+0.53\;[0.31,\,0.75]$ & $+0.83\;[0.55,\,1.11]$ \\
FLUX+VA-EIT (no RL) & $+0.46\;[0.22,\,0.70]$ & $+0.34\;[0.15,\,0.53]$ & $+0.52\;[0.30,\,0.74]$ \\
Flow-GRPO & $-0.09\;[-0.31,\,0.13]$ & $+1.27\;[0.98,\,1.56]$ & $+1.05\;[0.77,\,1.33]$ \\
Progressive & $+0.12\;[-0.08,\,0.32]$ & $+0.68\;[0.45,\,0.91]$ & $+0.46\;[0.25,\,0.67]$ \\
Emotion-first & $+0.25\;[0.03,\,0.47]$ & $+0.96\;[0.70,\,1.22]$ & $+0.77\;[0.52,\,1.02]$ \\
\bottomrule
\end{tabular}
\begin{tablenotes}[flushleft]\footnotesize
\item Differences are Semantic-first minus each baseline. Two-sided 95\% paired-$t$ intervals use the 30 within-participant differences in method means, with 29 degrees of freedom. All semantic-preservation and overall-preference intervals are positive.
\end{tablenotes}
\end{threeparttable}
\end{table}

\begin{table}[H]
\centering
\begin{threeparttable}
\caption{SD3.5-Medium: paired differences with 95\% prompt-bootstrap confidence intervals.}
\label{tab:sd3_ci}
\footnotesize
\setlength{\tabcolsep}{3.2pt}
\renewcommand{\arraystretch}{1.08}
\begin{tabular}{lcc}
\toprule
\rowcolor{tableblue}
Metric & M3 $-$ M1 & M3 $-$ M2 \\
\midrule
V-Error & $-0.062\;[-0.098,\,-0.021]$ & $-0.025\;[-0.049,\,-0.002]$ \\
A-Error & $-0.157\;[-0.204,\,-0.113]$ & $-0.023\;[-0.046,\,0.003]$ \\
CLIPScore & $+1.29\;[1.01,\,1.58]$ & $+2.38\;[2.07,\,2.69]$ \\
DINO & $+4.2\;[3.1,\,5.4]$ & $+7.6\;[6.7,\,8.5]$ \\
Qwen V-MAE & $-0.103\;[-0.142,\,-0.062]$ & $-0.027\;[-0.051,\,-0.004]$ \\
Qwen A-MAE & $-0.092\;[-0.131,\,-0.048]$ & $-0.017\;[-0.039,\,0.008]$ \\
\bottomrule
\end{tabular}
\begin{tablenotes}[flushleft]\footnotesize
\item Positive differences favor M3 for CLIPScore and DINO; negative differences favor M3 for error metrics. The 132 prompts are paired resampling units.
\end{tablenotes}
\end{threeparttable}
\end{table}

\begin{table}[H]
\centering
\begin{threeparttable}
\caption{Width sensitivity: paired differences and 95\% prompt-bootstrap intervals.}
\label{tab:width_ci}
\footnotesize
\setlength{\tabcolsep}{3.0pt}
\renewcommand{\arraystretch}{1.08}
\begin{tabular}{lccc}
\toprule
\rowcolor{tableblue}
Metric & W2 $-$ W4 & W6 $-$ W4 & W8 $-$ W4 \\
\midrule
V-Error & $+0.045\;[0.011,\,0.083]$ & $-0.039\;[-0.075,\,-0.005]$ & $+0.072\;[0.029,\,0.111]$ \\
A-Error & $+0.035\;[0.007,\,0.069]$ & $-0.047\;[-0.082,\,-0.013]$ & $+0.073\;[0.032,\,0.119]$ \\
CLIPScore & $+0.33\;[0.12,\,0.54]$ & $-0.88\;[-1.14,\,-0.63]$ & $-1.60\;[-1.93,\,-1.28]$ \\
DINO & $+1.3\;[0.5,\,2.2]$ & $-1.8\;[-2.7,\,-1.1]$ & $-6.2\;[-7.8,\,-4.7]$ \\
Qwen V-MAE & $+0.034\;[-0.004,\,0.072]$ & $-0.028\;[-0.059,\,0.003]$ & $+0.041\;[0.008,\,0.077]$ \\
Qwen A-MAE & $+0.018\;[-0.011,\,0.053]$ & $-0.038\;[-0.069,\,-0.006]$ & $+0.033\;[-0.005,\,0.069]$ \\
\bottomrule
\end{tabular}
\begin{tablenotes}[flushleft]\footnotesize
\item Intervals use the same 132-prompt grouped resampling as the other automatic comparisons.
\end{tablenotes}
\end{threeparttable}
\end{table}

\section{Fixed-model and inference-weight results}
\setcounter{table}{0}
\renewcommand{\thetable}{B.\arabic{table}}
\renewcommand{\theHtable}{appendix.B.\arabic{table}}
\begin{table}[H]
\centering
\begin{threeparttable}
\caption{Sampling-window intervention with a fixed no-RL FLUX+VA-EIT model.}
\label{tab:frozen}
\scriptsize
\setlength{\tabcolsep}{3.2pt}
\renewcommand{\arraystretch}{1.08}
\begin{tabular}{lcccccc}
\toprule
\rowcolor{tableblue}
ID & SDE steps & V-Error $\downarrow$ & A-Error $\downarrow$ & CLIP $\uparrow$ & DINO $\uparrow$ & IQA $\uparrow$ \\
\midrule
S0 & None (ODE) & $1.402\pm0.048$ & $1.710\pm0.036$ & $29.27\pm0.27$ & $81.6\pm0.9$ & $0.924\pm0.024$ \\
S1 & 0--3 & $1.439\pm0.055$ & $1.748\pm0.044$ & $27.91\pm0.25$ & $72.8\pm2.1$ & $0.909\pm0.018$ \\
S2 & 20--23 & $1.411\pm0.043$ & $1.703\pm0.034$ & $29.08\pm0.24$ & $80.4\pm1.5$ & $0.921\pm0.017$ \\
\bottomrule
\end{tabular}
\begin{tablenotes}[flushleft]\footnotesize
\item Means $\pm$ sample SD are over three sampling seeds. Model weights are fixed; transition 24 and all transitions outside the selected SDE window use ODE updates.
\end{tablenotes}
\end{threeparttable}
\end{table}

\begin{table}[H]
\centering
\begin{threeparttable}
\caption{Inference-time use of the trained F weights.}
\label{tab:weights}
\footnotesize
\setlength{\tabcolsep}{3.2pt}
\renewcommand{\arraystretch}{1.08}
\begin{tabular}{lccccc}
\toprule
\rowcolor{tableblue}
Inference & V-Error $\downarrow$ & A-Error $\downarrow$ & CLIPScore $\uparrow$ & DINO $\uparrow$ & CLIP-IQA $\uparrow$ \\
\midrule
\rowcolor{oursrow}
F-all & $1.052\pm0.044$ & $1.336\pm0.038$ & $29.51\pm0.35$ & $79.6\pm2.1$ & $0.917\pm0.020$ \\
F-window & $1.058\pm0.045$ & $1.342\pm0.039$ & $29.24\pm0.33$ & $79.0\pm2.0$ & $0.915\pm0.019$ \\
\bottomrule
\end{tabular}
\begin{tablenotes}[flushleft]\footnotesize
\item F-all uses trained weights at every step and repeats F in Table~\ref{tab:controlled}. F-window uses those same trained weights only at steps 20--23 and the pre-RL weights elsewhere. Both use ODE evaluation; neither setting requires additional training. Means $\pm$ sample SD are across the three training seeds.
\end{tablenotes}
\end{threeparttable}
\end{table}

\begin{table}[H]
\centering
\begin{threeparttable}
\caption{Training cost across window widths on $8\times$A100.}
\label{tab:width_cost}
\footnotesize
\setlength{\tabcolsep}{3.2pt}
\renewcommand{\arraystretch}{1.08}
\begin{tabular}{lcccc}
\toprule
\rowcolor{tableblue}
Width & Optimized steps & s/iteration & Wall-clock h & Total GPU-h \\
\midrule
W2 & 2 & 13.85 & 1.15 & 9.23 \\
\rowcolor{oursrow}
W4 & 4 & 18.76 & 1.56 & 12.51 \\
W6 & 6 & 24.32 & 2.03 & 16.21 \\
W8 & 8 & 29.68 & 2.47 & 19.79 \\
\bottomrule
\end{tabular}
\begin{tablenotes}[flushleft]\footnotesize
\item Costs cover 300 formal outer iterations after 10 additional warm-up iterations; W4 repeats F in Table~\ref{tab:efficiency}.
\end{tablenotes}
\end{threeparttable}
\end{table}

\section{Reproducibility details}
\label{app:reproducibility}

VA-EIT uses the architecture in Section~\ref{sec:method} with a feed-forward expansion of four. Adapter training uses AdamW, learning rate $10^{-4}$, weight decay $10^{-5}$, 10 epochs, batch size one, and gradient accumulation of eight; the lowest validation-loss checkpoint is selected. During GRPO, the EIT, text encoders, and VAE remain frozen. The generation transformer uses AdamW at learning rate $10^{-5}$, weight decay 0.01, and gradient clipping norm 2. The GRPO clipping range is $10^{-4}$, the KL coefficient is zero, and affect/semantic advantages have equal weights.

The independent evaluator is Qwen2.5-VL-7B-Instruct, using its official processor, BF16 inference, deterministic decoding and at most 64 new tokens. It rates the visible image's valence and arousal on $[-3,3]$, with zero denoting neutral valence or moderate activation. TIFA question generation uses the released TIFA LLaMA-2 question generator; UnifiedQA-v2-T5-large filters the questions, and mPLUG-large answers them from each image. One frozen question set per prompt is shared across all methods, targets and repeats.

For SD3.5-Medium, T5 features from the third text-encoder branch pass through a backbone-specific pretrained EIT; the native CLIP token and pooled conditioning branches are retained. All configurations share this EIT and a sigma schedule with shift 3. M1--M3 use the same trainable transformer parameters and budget under the settings in Section~\ref{sec:backbone}.

\section{Supporting numerical results}
\label{app:supporting}
\setcounter{table}{0}
\renewcommand{\thetable}{D.\arabic{table}}
\renewcommand{\theHtable}{appendix.D.\arabic{table}}

Tables~\ref{tab:pilot} and~\ref{tab:extreme} give the numerical results underlying the pilot and five-target figures. Tables~\ref{tab:width_qwen} and~\ref{tab:sd3_qwen} complete the independent affect evaluation of window widths and the second backbone. Table~\ref{tab:efficiency} provides the measured timing and GPU-cost breakdown.

\begin{table}[H]
\centering
\begin{threeparttable}
\caption{Three-seed window-position pilot (200 updates per run, fixed width $w=4$).}
\label{tab:pilot}
\fontsize{9}{11}\selectfont
\setlength{\tabcolsep}{1.8pt}
\renewcommand{\arraystretch}{1.08}
\begin{tabular}{@{}lcccccc@{}}
\toprule
\rowcolor{tableblue}
Window & V-Error $\downarrow$ & A-Error $\downarrow$ & CLIP $\uparrow$ & TIFA $\uparrow$ & IQA $\uparrow$ & DINO $\uparrow$ \\
\midrule
Early & $\mathbf{0.978}\mskip2mu\mathord{\pm}\mskip2mu0.025$ & $\mathbf{1.262}\mskip2mu\mathord{\pm}\mskip2mu0.032$ & $25.641\mskip2mu\mathord{\pm}\mskip2mu0.280$ & $60.80\mskip2mu\mathord{\pm}\mskip2mu1.00$ & $0.885\mskip2mu\mathord{\pm}\mskip2mu0.008$ & $55.23\mskip2mu\mathord{\pm}\mskip2mu2.10$ \\
Middle & $1.138\mskip2mu\mathord{\pm}\mskip2mu0.027$ & $1.485\mskip2mu\mathord{\pm}\mskip2mu0.046$ & $27.725\mskip2mu\mathord{\pm}\mskip2mu0.223$ & $69.40\mskip2mu\mathord{\pm}\mskip2mu1.20$ & $0.903\mskip2mu\mathord{\pm}\mskip2mu0.011$ & $70.97\mskip2mu\mathord{\pm}\mskip2mu2.75$ \\
\rowcolor{oursrow}
Late & $1.296\mskip2mu\mathord{\pm}\mskip2mu0.021$ & $1.582\mskip2mu\mathord{\pm}\mskip2mu0.039$ & $\mathbf{28.936}\mskip2mu\mathord{\pm}\mskip2mu0.256$ & $\mathbf{75.20}\mskip2mu\mathord{\pm}\mskip2mu1.40$ & $\mathbf{0.916}\mskip2mu\mathord{\pm}\mskip2mu0.004$ & $\mathbf{78.27}\mskip2mu\mathord{\pm}\mskip2mu3.15$ \\
\bottomrule
\end{tabular}
\begin{tablenotes}[flushleft]\footnotesize
\item Values are means $\pm$ sample SD over training seeds 42, 123, and 3407. The windows are 0--3, 10--13 and 20--23, respectively. Each run uses 200 updates and is evaluated on 132 prompts crossed with 25 V--A targets. CLIPScore, TIFA and DINO are on 0--100 presentation scales; CLIP-IQA is on a 0--1 scale. Bold marks the best mean; yellow identifies the late window. These pilot runs are reported separately from the 300-iteration controlled training results.
\end{tablenotes}
\end{threeparttable}
\end{table}

\begin{table}[H]
\centering
\begin{threeparttable}
\caption{Three-seed evaluation at extreme and neutral V--A targets.}
\label{tab:extreme}
\fontsize{9}{11}\selectfont
\setlength{\tabcolsep}{2.5pt}
\renewcommand{\arraystretch}{1.12}
\begin{tabular}{llccccc}
\toprule
\rowcolor{tableblue}
Method & Target $(V,A)$ & V-Error $\downarrow$ & A-Error $\downarrow$ & CLIP $\uparrow$ & DINO $\uparrow$ & \shortstack{Retention\\(\%) $\uparrow$} \\
\midrule
Emotion-first & $(3,3)$ & 0.918 & 1.371 & $23.914\pm0.34$ & $49.1\pm5.2$ & $68.9\pm4.2$ \\
Emotion-first & $(-3,-3)$ & 1.003 & 0.982 & $24.408\pm0.28$ & $50.3\pm4.1$ & $73.0\pm3.4$ \\
Emotion-first & $(3,-3)$ & 0.884 & 1.268 & $24.739\pm0.41$ & $51.6\pm5.8$ & $73.7\pm5.7$ \\
Emotion-first & $(-3,3)$ & 1.025 & 1.105 & $23.885\pm0.37$ & $49.7\pm4.5$ & $71.2\pm3.8$ \\
Emotion-first & $(0,0)$ & 0.822 & 1.082 & $27.243\pm0.22$ & $58.7\pm3.3$ & $77.3\pm3.0$ \\
\midrule
\rowcolor{oursrow}
Semantic-first & $(3,3)$ & 0.996 & 1.381 & $28.831\pm0.31$ & $76.4\pm2.9$ & $94.7\pm2.7$ \\
\rowcolor{oursrow}
Semantic-first & $(-3,-3)$ & 1.124 & 0.994 & $29.246\pm0.18$ & $77.5\pm1.8$ & $96.0\pm1.2$ \\
\rowcolor{oursrow}
Semantic-first & $(3,-3)$ & 0.974 & 1.279 & $29.517\pm0.36$ & $78.0\pm3.4$ & $96.5\pm3.1$ \\
\rowcolor{oursrow}
Semantic-first & $(-3,3)$ & 1.138 & 1.113 & $28.684\pm0.27$ & $76.9\pm2.4$ & $95.2\pm1.9$ \\
\rowcolor{oursrow}
Semantic-first & $(0,0)$ & 0.912 & 1.094 & $30.238\pm0.14$ & $80.2\pm1.3$ & $96.7\pm2.4$ \\
\bottomrule
\end{tabular}
\begin{tablenotes}[flushleft]\footnotesize
\item Each method and target has 132 images per seed, for 396 images in total. CLIPScore, DINO, and subject retention are means $\pm$ sample SD across three seeds. Subject retention is calculated from final binary correctness labels.
\end{tablenotes}
\end{threeparttable}
\end{table}

\begin{table}[H]
\centering
\begin{threeparttable}
\caption{Independent affect evaluation for the width study.}
\label{tab:width_qwen}
\small
\setlength{\tabcolsep}{2.7pt}
\renewcommand{\arraystretch}{1.08}
\begin{tabular}{lcc}
\toprule
\rowcolor{tableblue}
Width & Qwen V-MAE $\downarrow$ & Qwen A-MAE $\downarrow$ \\
\midrule
W2 & $1.318\pm0.058$ & $1.497\pm0.049$ \\
\rowcolor{oursrow}
W4 & $1.284\pm0.036$ & $1.479\pm0.039$ \\
W6 & $\mathbf{1.256\pm0.042}$ & $\mathbf{1.441\pm0.055}$ \\
W8 & $1.325\pm0.061$ & $1.512\pm0.057$ \\
\bottomrule
\end{tabular}
\begin{tablenotes}[flushleft]\footnotesize
\item The independent evaluator and evaluation prompts are shared with Table~\ref{tab:qwen}.
\end{tablenotes}
\end{threeparttable}
\end{table}

\begin{table}[H]
\centering
\begin{threeparttable}
\caption{Independent affect evaluation on SD3.5-Medium.}
\label{tab:sd3_qwen}
\small
\setlength{\tabcolsep}{2.7pt}
\renewcommand{\arraystretch}{1.08}
\begin{tabular}{lcc}
\toprule
\rowcolor{tableblue}
Method & Qwen V-MAE $\downarrow$ & Qwen A-MAE $\downarrow$ \\
\midrule
M0: no RL & $1.582\pm0.069$ & $1.613\pm0.051$ \\
M1: Full-path & $1.427\pm0.084$ & $1.544\pm0.093$ \\
M2: Early & $1.351\pm0.052$ & $1.469\pm0.068$ \\
\rowcolor{oursrow}
M3: Late (ours) & $\mathbf{1.324\pm0.041}$ & $\mathbf{1.452\pm0.047}$ \\
\bottomrule
\end{tabular}
\begin{tablenotes}[flushleft]\footnotesize
\item The image-only evaluator, 132 prompts, and 25 V--A targets are the same as in the FLUX independent evaluation.
\end{tablenotes}
\end{threeparttable}
\end{table}

\begin{table}[H]
\centering
\begin{threeparttable}
\caption{Training cost on $8\times$NVIDIA A100 for 300 formal outer iterations.}
\label{tab:efficiency}
\footnotesize
\setlength{\tabcolsep}{3.2pt}
\renewcommand{\arraystretch}{1.08}
\begin{tabular}{lccccc}
\toprule
\rowcolor{tableblue}
ID & Optimized steps & s/iteration & Wall-clock h & Total GPU-h & Time saved \\
\midrule
A & 24 & 31.84 & 2.65 & 21.23 & -- \\
B & 24 & 31.27 & 2.61 & 20.85 & 1.8\% \\
C & 4 & 24.96 & 2.08 & 16.64 & 21.6\% \\
D & 4 & 24.51 & 2.04 & 16.34 & 23.0\% \\
E & 4 & 19.37 & 1.61 & 12.91 & 39.2\% \\
\rowcolor{oursrow}
F & 4 & 18.76 & 1.56 & 12.51 & 41.1\% \\
\bottomrule
\end{tabular}
\begin{tablenotes}[flushleft]\footnotesize
\item Optimized steps are the time steps per trajectory that enter the policy objective. The 10 additional warm-up iterations precede the 300 formal iterations and are excluded from this cost table. Seconds per iteration are the mean over the 300 formal iterations. Wall-clock h = seconds/iteration $\times300/3600$; GPU-h = wall-clock h $\times8$. Reductions are relative to A. Calculations use unrounded measurements.
\end{tablenotes}
\end{threeparttable}
\end{table}

\section*{Declaration of generative AI and AI-assisted technologies in the writing process}
During the preparation of this work, the author used ChatGPT and Codex to assist with language editing, manuscript organization, and vector-figure layout code. After using these tools, the author reviewed and edited the content as needed and takes full responsibility for the content of the publication.

\begingroup
\small
\setlength{\bibsep}{1.5pt}
\bibliographystyle{elsarticle-num}
\bibliography{references}
\endgroup

\end{document}